\documentclass[conference]{IEEEtran}
\IEEEoverridecommandlockouts
\usepackage{cite}
\usepackage{amsmath,amssymb,amsfonts}
\usepackage{graphicx}
\usepackage{textcomp}
\usepackage{algorithm}
\usepackage{algpseudocode}
\usepackage[table,xcdraw]{xcolor}
\usepackage{subcaption}
\def\BibTeX{{\rm B\kern-.05em{\sc i\kern-.025em b}\kern-.08em
    T\kern-.1667em\lower.7ex\hbox{E}\kern-.125emX}}
\begin{document}

\title{Leveraging Pre-Images to Discover Nonlinear Relationships in Multivariate Environments\\
\thanks{This research was partly funded by an Ernest J. Del Monte Institute for Neuroscience Award from the Harry T. Mangurian Jr. Foundation. This work was conducted as a Practice Quality Improvement (PQI) project related to American Board of Radiology (ABR) Maintenance of Certificate (MOC) for Prof. Dr. Axel Wismüller. We thank Dr. John Foxe, Institute for Neuroscience, University of Rochester, for his support.}
}

\author{\IEEEauthorblockN{M. Ali Vosoughi\\ \textit{Graduate Student Member, IEEE}}
\IEEEauthorblockA{{Dept. of Electrical and Comp. Eng.}\\
{{University of Rochester, Rochester, NY, USA}}\\
{Email: mvosough@ece.rochester.edu}}
\and
\IEEEauthorblockN{Axel Wismüller$^{1, 2, 3, 4}$}
\IEEEauthorblockA{{$^{1}$Dept. of Imaging Sciences, $^{2}$Electrical and Comp. Eng., $^{3}$Biomed. Eng.,}\\
{University of Rochester Medical Center, Rochester, NY, USA}\\
{$^{4}$Faculty of Medicine, Ludwig Maximilian University, Munich, Germany}
}
}

\maketitle

\begin{abstract}

Causal discovery, beyond the inference of a network as a collection of connected dots, offers a crucial functionality in scientific discovery using artificial intelligence. The questions that arise in multiple domains, such as physics, physiology, the strategic decision in uncertain environments with multiple agents, climatology, among many others, have roots in causality and reasoning. It became apparent that many real-world temporal observations are nonlinearly related to each other. While the number of observations can be as high as millions of points, the number of temporal samples can be minimal due to ethical or practical reasons, leading to the curse-of-dimensionality in large-scale systems. This paper proposes a novel method using kernel principal component analysis and pre-images to obtain nonlinear dependencies of multivariate time-series data. We show that our method outperforms state-of-the-art causal discovery methods when the observations are restricted by time and are nonlinearly related. Extensive simulations on both real-world and synthetic datasets with various topologies are provided to evaluate our proposed methods.

\end{abstract}

\section{INTRODUCTION}
Causal relationships are meaningful and apply to many problems, as evidenced by the DARPA's recent challenge on machine common-sense (MCS) \cite{darpa_MCS}. Such applications require an AI machine with a mechanism to understand and reason the system's underlying relations, which solo statistics cannot plead (see \cite{shpitser2008complete} for the condition that statistics fail). One may argue that such AI machines are causal because AI that finds rules is equivalent to AI that finds causality: when a causal relationship is strong enough, we call it a rule \cite[Chapter~1]{pearl2018book}. An inherent advantage of learning reasons is learning with few samples. Causality gained the most traction recently to build models with sufficient robustness, generalizability, interpretability, and the capability to discover semantically meaningful representations. 

The completeness of identifiability and development of \textit{do-calculus} paved the way for causal inference using observational data \cite{shpitser2008complete}. However, using solo temporal measurements, the generative process underneath data is unknown, making the use of \textit{do-calculus} unsettled. Granger causality (GC) is widely used to infer causality in the time-series data given the faithfulness assumption. GC initially stated for linear relations \cite{granger1988some}, and states that a time-series A is a cause for time-series B if A causes a better prediction for B.

Although there is a rich body of literature on discovering nonlinear relations from time-series data in the recent decade, yet it is a challenging problem for two reasons: 1) the detected nonlinear relations can be spurious, which requires a multivariate approach, and 2) the curse-of-dimensionality and ill-posedness becomes the main problem with an increasing number of nodes \cite{runge2019detecting, antonacci2021testing}. Various algorithms have been proposed to recover nonlinear relations in networks, including Ancona's work \cite{ancona2004radial}, where they present conditions to detect nonlinear relationships, and in \cite{marinazzo2008kernel} Marinazzo develops a method, called kernel Granger causality (KGC) to consider spurious relations using kernel functions. Unfortunately, their approach in \cite{ancona2004radial} is bivariate and cannot be extended to multivariate settings, and \cite{marinazzo2008kernel} cannot be applied to large-scale problems due to over-fitting. In parallel with the Marinazzo approach, another method called transfer entropy (TE) was proposed by Schreiber \cite{schreiber2000measuring}, which formulates the causal relationship among time-series from the information theory's perspective. In his 2009 paper, Barnett proved that the TE and GC were equivalent for Gaussian variables \cite{barnett2009granger}. Therefore, the KGC under the Gaussian assumption is equal to TE. These two lines of work (TE and KGC) have received considerable attention.

Several algorithms have been proposed to extended the previously discussed approaches to multivariate time-series \cite{kraskov2004estimating, wollstadt2018idtxl, runge2019detecting, marinazzo2011nonlinear, runge2019inferring, vosoughi2021marijuana}; however, the curse-of-dimensionality of the large-scale problem yet is an open problem \cite{runge2019detecting}. Note that in real-world studies such as climatology, the dimensions of the problem can reach millions of variables \cite{camps2019perspective}, so in the presence of short time-series (just a few decades of temperature measurement), the inverse problems quickly become ill-posed due to the presence of redundant variables. In large-scale problems, TE-based methods are costly based on their dependency on estimating the probability distribution functions. Our analysis of a recent TE algorithm based on the Kraskov estimation \cite{wollstadt2018idtxl} for a graph with 34 nodes takes 26 hours on an Nvidia GeForce 1080-Ti GPU and tens of times more on a CPU, while the cost increases drastically with the number of nodes. 

The closest study to our work in terms of the ongoing open problem is the PCMCI method \cite{runge2019detecting}. Peter and Clark Momentary Conditional Independence (PCMCI) is a two-step algorithm, with an statistical independence test in the PC stage, and obtaining causality strengths of the significant links in the MCI stage \cite{runge2019detecting}. Besides, because our method is a nonlinear Granger causality under Gaussian assumption, it is equivalent to the TE \cite{wollstadt2018idtxl, kraskov2004estimating, barnett2009granger} for Gaussian variables. However, our proposed method differs from the above methods in two ways. First, since it is a nonlinear GC method, it is equivalent to the TE method for Gaussian variables, while it is also suitable for large-scale problems. On the other hand, although PCMCI is proposed for large-scale problems, the method we present provides the ability to apply to short time-series on a large scale and provides all steps without examining nonlinear independencies using a partial correlation as is in the PCMCI. We believe our contribution adds numerous capabilities to the existing literature, affirmed through extensive simulations on synthetic and real-world datasets.

\section{PRELIMINARIES} \label{sec:prems}
In an abuse of notation, we use $i, j$ and $Y_i, Y_j$ to point to the nodes and random variables on the nodes. Scalars are small letters, vectors are in small bold letters, and matrices are in bold capital letters. Calligraphic letters $\mathcal{N}, \mathcal{E}, \mathcal{H}$, and $\mathcal{Y}$ represent the set of nodes, set of weighted edges, feature space (Hilbert subspace), and input space, correspondingly. 
\paragraph{Causality assumptions}
A variable $i$ is said to be a \textit{cause} of a variable $j$ if $j$ can change in response to change in $i$. Consider a \textit{directional graphical model} $G=(\mathcal{N},\mathcal{E})$ consisting of the set of nodes $\mathcal{N}=\{ i|i=1,\ldots,N\}$ and the set of directed edges $\mathcal{E}$, whose topology is unknown, but time-series $\{y_{it}\}^T_{t=1}$ are observed per node $i$, over $T$ time intervals. We assume that \textit{edges are causal}, meaning that every parent $i$ is a direct cause of all its children $\mathcal{C}(i)$, and therefore $i$ is a parent of $j$, denoting as $i\in\mathcal{P}(j)$. Furthermore, we assume that measurements $\{y_{it}\}^T_{t=1}$ are \textit{causally sufficient}, meaning that there are no unobserved confounders of any of the variables in the graph\footnote{Note that we cannot test unconfoundedness, and cannot guarantee that it is satisfied. In other words, no unobserved confounding is unrealistic assumption, and we will always have unobserved confounders \cite{manski2003partial}}. Our goal is to estimate causal quantities $\delta_{ij}$ between each pair of the nodes $i$ and $j$ in the presence of all other nodes ${Z}=\mathcal{N}\setminus \{i,j\}$, and derive the topology of the graph as  $\mathcal{E}=\{\delta_{ij}| i,j\in \mathcal{N}\}$, such that $i \in \mathcal{P}(j)\text{ if }\delta_{ij}>0 \text{ else } i \not\in \mathcal{P}(j)$. We assume that $\delta_{ij}$ is \textit{identifiable}, therefore we can compute it from a purely statistical quantity\footnote{Identifiablity requires to have four assumptions: unconfoundness, positivity, consistency, and no interference.}\cite{shpitser2008complete}. We use \textit{faithfulness} assumption to obtain causal relations among the nodes $\mathcal{N}$, which allows us to infer \textit{d-separations} in the graph from in dependencies in the distributions $Y_i  {\bot \!\!\!\bot }_{G} Y_j \vert {Z} \Leftarrow Y_i {\bot \!\!\!\bot }_P Y_j \vert {Z}$. The notion ${\bot \!\!\!\bot }$ implies independence in the graph $G$, or the distribution $P(Y_1,\ldots, Y_N)$ where $Y_i$ denotes the random variable corresponding to node $i$. Note that faithfulness is a weak assumption; however, without it, we cannot infer the topology of the causal graph from the observational time-series\footnote{The reason is related to global Markov assumption, where given that P is Markov with respect to graph $G$ one can use $G$ to infer independencies in $P$. However, in case of the unstructured time-series data, the graph topology is priory unknown and we cannot use Markov assumption before obtaining the graph topology.}.
\paragraph{Causal quantities}
Linear vector autoregressive model (VARM) is widely adopted framework to infer Granger causality in multivariate time-series, in which a \textit{Granger causality index} $\delta_{ij}$ can be defined as a quantification of prediction quality as follows. VARM states that each observed $\mathbf{y}_{t}:=[{y}_{1t},\ldots,{y}_{Nt}]^\top$ is a linear combination of the time-lagged versions of the measurements $\{\{y_{i(t-\ell )}\}^N_{i=1}\}^L_{\ell=1}$. Let $\mathbf{A}^{\ell}\in\mathbb{R}^{N\times N}$ denote the “time-lagged” VARM parameters matrix, with $[\mathbf{A}^{\ell}]_{ij}=a_{ij}^{\ell}$, and $a_{ij}^{\ell}$ as model coefficients over a lag of $\ell$ time points. Given the time-lagged multivariate time-series $\{\mathbf{y}_{(t-\ell)}\}^L_{t=\ell}$, where , the goal is to estimate the model parameter matrices $\{\mathbf{A}^{\ell}\}_{\ell=1}^L$ in:
\begin{equation} 
\mathbf{y}_t=\sum_{\ell=1}^{L}\mathbf{A}^{\ell}\mathbf{y}_{(t-\ell)}+\mathbf{e}_{t},\label{eqn:linear_autoregressive_model}
\end{equation}
and minimize the residual errors $\mathbf{e}_{t}$ using an optimization method, such as ordinary least squares (OLS), the lasso, the ridge, or the elastic-net regressions \cite{friedman2010regularization}. 

Let $\mathbf{E}:=[\mathbf{e}_\ell,\ldots,\mathbf{e}_T]$ be the matrix of residuals of VARM for the full system including all nodes $\mathcal{N}$, and let $\mathbf{E}^{i-}:=[\mathbf{e}^{i-}_\ell,\ldots,\mathbf{e}^{i-}_T]$ to be the same matrix when the node $i$ is excluded $\mathcal{N}\setminus \{i\}$ from the VARM, where each $\mathbf{e}^{i-}_t$ is obtained using $\mathbf{y}^{i-}_t:=[{y}_{1t},\dots, {y}_{(i-1)t}, {y}_{(i+1)t},\dots,{y}_{Nt}]^\top$ in expression \eqref{eqn:linear_autoregressive_model}. Then, the derivation of error covariance matrices $\Sigma=cov(\mathbf{E},\mathbf{E})$ and $\Sigma^{i-}=cov(\mathbf{E}^{i-},\mathbf{E}^{i-})$ will be straightforward. Based on $\Sigma$ of the full VARM and $\Sigma^{i-}$ of the VARM without $i$, the degree of information flow from node $i$ to node $j$ can be quantified by $\ln({\Sigma_{j}^{i-}}/{\Sigma_{j}})$, where $\Sigma_{j}^{i-}$ and $\Sigma_{j}$ denote the diagonal entries of $\Sigma^{i-}$ and $\Sigma$ associated to node $j$, respectively.  Consequently, the topology of the causal graph can be inferred using:

\begin{equation} \delta_{ij}=\left\{ {
\begin{array}{lll} 

\ln({\Sigma_{j}^{i-}}/{\Sigma_{j}}), & &\ln({\Sigma_{j}^{i-}}/{\Sigma_{j}})>0 \\
0, & & \ln({\Sigma_{j}^{i-}}/{\Sigma_{j}})\le 0 ,
\end{array}} \right.
\end{equation}
as $\mathcal{E}=\{\delta_{ij}| i,j\in \mathcal{N}, i \in \mathcal{P}(j)\text{ if }\delta_{ij}>0 \text{ else } i \not\in \mathcal{P}(j)\}$.

\paragraph{Problem Statement}  
Given nodal measurements in \eqref{eqn:linear_autoregressive_model}, the aim is to solve VARM in the reproducing kernel Hilbert space (RKHS) and then revert the solution (residuals) to the input space to infer the topology in the input space. The benefit is to encompass nonlinearities while reducing dimensionality to address the curse-of-dimensionality.
\section{The proposed method}\label{sec:method}
\begin{figure}
    \centering
    \includegraphics[width=0.9\linewidth]{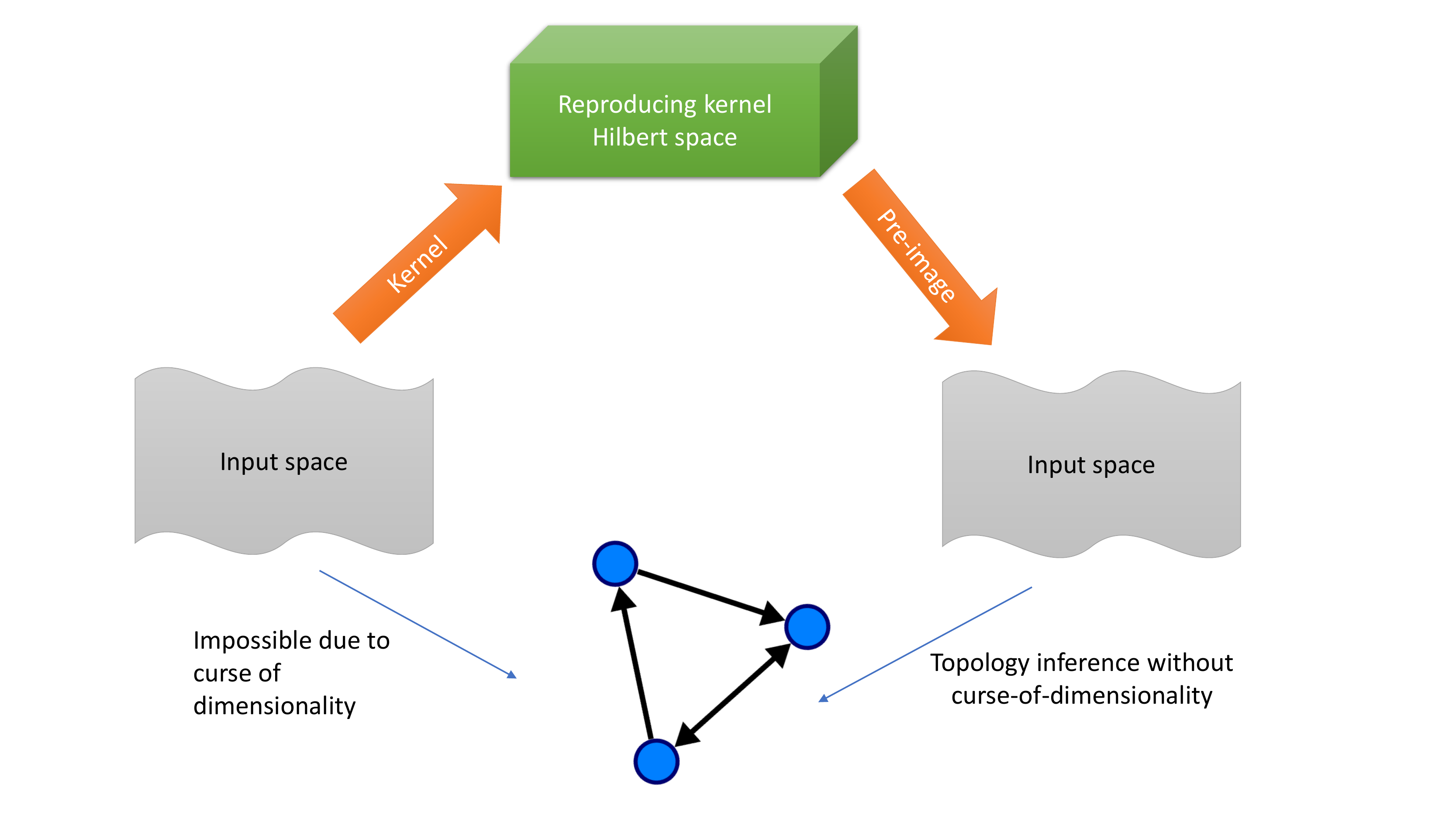}
    \caption{The proposed method involves three steps: kernel PCA, solving linear vector autoregressive model (VARM) in feature space, and using pre-images to infer topology in input space.}
    \label{fig:proposed_method}
\end{figure}

The proposed method consists of three steps using three off-the-shelf modules, as shown in Fig. \ref{fig:proposed_method}. The first step involves a transformation of the input space to the dimensionality reduced (or lifted) feature space using kernel principal component (kernel PCA), the second step involves solving the VARM of \eqref{eqn:linear_autoregressive_model} in the feature space, and the third step which is the essential part of the method, is to revert the residuals of the VARM in the feature space into the input space using pre-images, and finally infer the topology of the graph $\mathcal{E}$ in the input space.
\paragraph{Dimensionality reduction using kernel PCA}
The aim is to transfer the equation \eqref{eqn:linear_autoregressive_model} to the space of the principal components with lower dimensions, encompassing nonlinear dependencies. 
We define RKHS function $\phi(.)$ as $\phi({y}_{jt}): \mathcal{Y}\rightarrow \mathcal{H}$ to project time-series measurements into the feature space.
A feature space related to the principal components with the maximum explained variance of data would be desired; therefore, kernel PCA is of interest.
Altough $P<N$ as the dimension of the feature space is desired, $ N \le P<\infty$ can be also used to to lift the space (rather than reducing). Therefore, $\{\phi_{p}(.)\}^P_{p=1}$ builds the basis of feature space, and $\mathbf{\phi}(y_{it}) :=[\phi_{1}(y_{it}), \ldots, \phi_{P}(y_{it})]^\top$. A projection of input data ${y}_{jt}$ onto the $p$ -th principal component  would be desired, such that the major proportion of data variation is explained by a few principal components \cite{scholkopf2002learning}. 
%


\paragraph{Solving the VARM in the reproducing kernel Hilbert space}
Using the RKHS function transformation, the feature space representation of \eqref{eqn:linear_autoregressive_model} will be:
\begin{equation} 
\eta_{pt}=\sum_{\ell=1}^{L}\sum_{j=1}^{N}\alpha_{jp}^{\ell}\phi_{p}({y}_{j(t-\ell)})+\varepsilon_{pt} \label{eqn:linear_autoregressive_feature_space}
\end{equation}
where $\{\eta_{pt}\}^T_{t=1}$ for $p=1,\ldots,P$ is the transformed time-series into the feature space, and $\{\varepsilon_{pt}\}^P_{p=1}$ is the  error at time slot $t$ in the feature space.
The expression \eqref{eqn:linear_autoregressive_feature_space} is a linear VARM and can be derived with the same approach as discussed in section \ref{sec:prems}.
Once the $\{\eta_{pt}\}^P_{p=1}$ in \eqref{eqn:linear_autoregressive_feature_space} are obtained , the predictions $\{\hat{\eta}_{pt}\}^P_{p=1}$ will be transformed into the input space to infer the graph topology.

\paragraph{Estimating pre-images}
Let $\mathbf{Y}:=[{\mathbf{y}}_1,\ldots,{\mathbf{y}}_T]$, $\mathbf{\gamma}(\eta_{pt}) :=[\gamma_{1}(\eta_{pt}), \ldots, \gamma_{N}(\eta_{pt})]^\top$, $\mathbf{\Gamma}_p := [\gamma(\eta_{p1})$, $ \ldots, $ $\gamma(\eta_{pT})]^\top$,
$\mathbf{\Gamma}:=[\mathbf{\Gamma}_1,\ldots,\mathbf{\Gamma}_P]$, $\mathbf{\Phi}_j:=[\mathbf{\phi}(y_{j1}),\ldots,\mathbf{\phi}(y_{jT})]^\top$, and $\mathbf{\Phi}:=[\mathbf{\Phi}_1,\ldots,\mathbf{\Phi}_P]$, and the notations $\mathbf{Y^{j-}}:=[\Tilde{\mathbf{y}}_1,\ldots,\Tilde{\mathbf{y}}_{j-1},\Tilde{\mathbf{y}}_{j+1},\ldots,\Tilde{\mathbf{y}}_N]$, $\Tilde{\mathbf{\eta}}_{p} :=[\eta_{p1},\ldots,\eta_{pT}]^\top$,  $\mathbf{H}:=[\Tilde{\mathbf{\eta}}_{1},\ldots,\Tilde{\mathbf{\eta}}_{P}]$  for the sake of simplicity.
To transform $\{\hat{\eta}_{pt}\}^P_{p=1}$ into the input space and obtain the pre-image ${\gamma}(\hat{\eta}_{pt}):\mathcal{H}\rightarrow\mathcal{Y}$, one needs to estimate $\hat{y}_{jt} = \arg \min \limits _{\hat{y}_{jt}\in \mathcal{Y}} \Vert{y}_{jt}-\gamma(\phi(\hat{y}_{jt}))\Vert^2$, which is a nonlinear optimization problem, and is not interesting as a solution to find pre-images \cite{scholkopf2002learning, bakir2003learning}. However, interestingly, both the inputs $\{y_{it}\}^T_{t=1}$ and their corresponding mappings are present for $\phi({y}_{jt}): \mathcal{Y}\rightarrow \mathcal{H}$, easing the pre-image problem to a simple learning problem. Therefore, we use the learning problem $\mathcal{L} = \arg \min \limits _{\mathbf{\Gamma}}  \Vert\mathbf{Y}- \mathbf{\Gamma}\mathbf{\Phi}^\top\Vert^2$ instead, which is adopted from \cite{bakir2003learning} to directly obtain coefficients  $\gamma_{jp}:=[\mathbf{\Gamma}]_{jp}$, and consequently the predictions $\hat{y}_{jt}=\sum^P_{p=1}\gamma_{jp} \hat{\eta}_{pt}$ will be obtained. Simultaneously, the topology of the graph can be obtained as discussed in section \ref{sec:prems}. 
The complete algorithm is shown in Algorithm. \ref{alg:complete_alg_kernel_lsGC}. Note that novel statistical frameworks such as \cite{mohammadi2021finite, mohammadi2021ultrasound, mohammadi2020statistical, mohammadi2021sspw} can be used to formulate the problem in a neural network, which can be as a future work. 
\begin{algorithm}
    \caption{The proposed method leveraging pre-images} \label{alg:complete_alg_kernel_lsGC}
    \begin{algorithmic}
        \Require $L,P,\mathbf{Y},\mathbf{\phi}(.)$
        \Ensure $\mathcal{E}\leftarrow \{\ln(\frac{\mathbf{\Sigma}_{j}^{i-}}{\mathbf{\Sigma}_{j}})\}_{j=1}^N
        \leftarrow \ln(\frac{diag\{cov(\mathbf{Y}^{i-}-\hat{\mathbf{Y}}^{i-})\}}{diag\{cov(\mathbf{Y}-\hat{\mathbf{Y}})\}})$
        \State  $\mathbf{\Gamma} \leftarrow  \mathcal{L}  \leftarrow \mathbf{Y} \leftarrow \text{normalize}(\mathbf{Y})$
        \For {$j=1$ to $N$} 
            \State $\mathbf{H}\xleftarrow[]{\mathbf{\Phi}} \mathbf{Y}$, and $\mathbf{H}^{j-}\xleftarrow[]{\mathbf{\Phi}} \mathbf{Y}^{j-}$
            \For{$\ell=1$ to $L$} 
            $\hat{\mathbf{Y}} \xleftarrow[]{\mathbf{\Gamma}} \hat{\mathbf{H}}$  and $\hat{\mathbf{Y}}^{j-} \xleftarrow[]{\mathbf{\Gamma}} \hat{\mathbf{H}}^{j-}$
            \EndFor
        \EndFor

    \end{algorithmic}

\end{algorithm}

\begin{figure}
\begin{subfigure}[b]{0.15\textwidth}
    \centering
    \includegraphics[width=\textwidth]{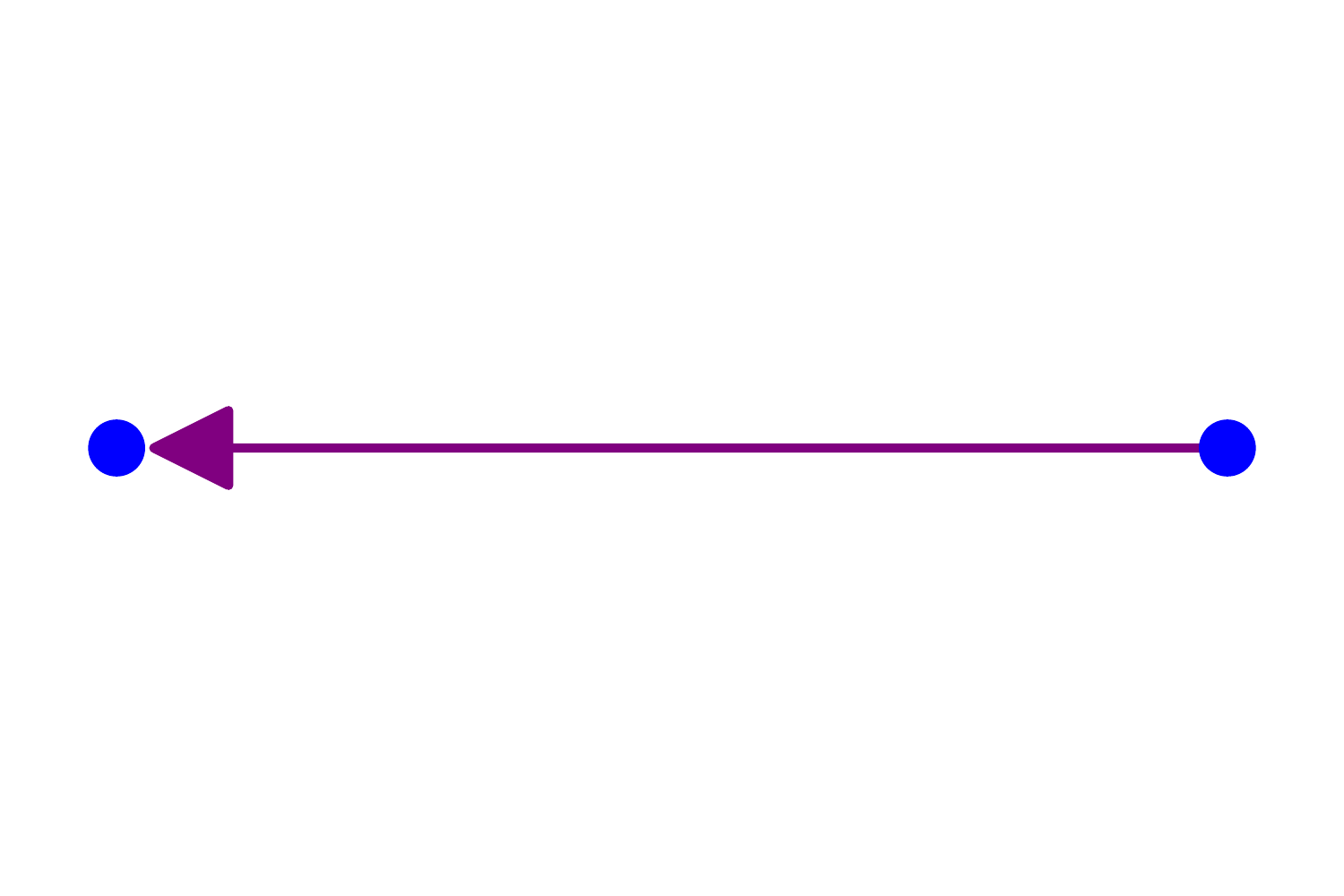}
    \caption{2- Logistic}
    \label{fig:logistic1}
\end{subfigure}
\begin{subfigure}[b]{0.15\textwidth}
    \centering
    \includegraphics[width=\textwidth]{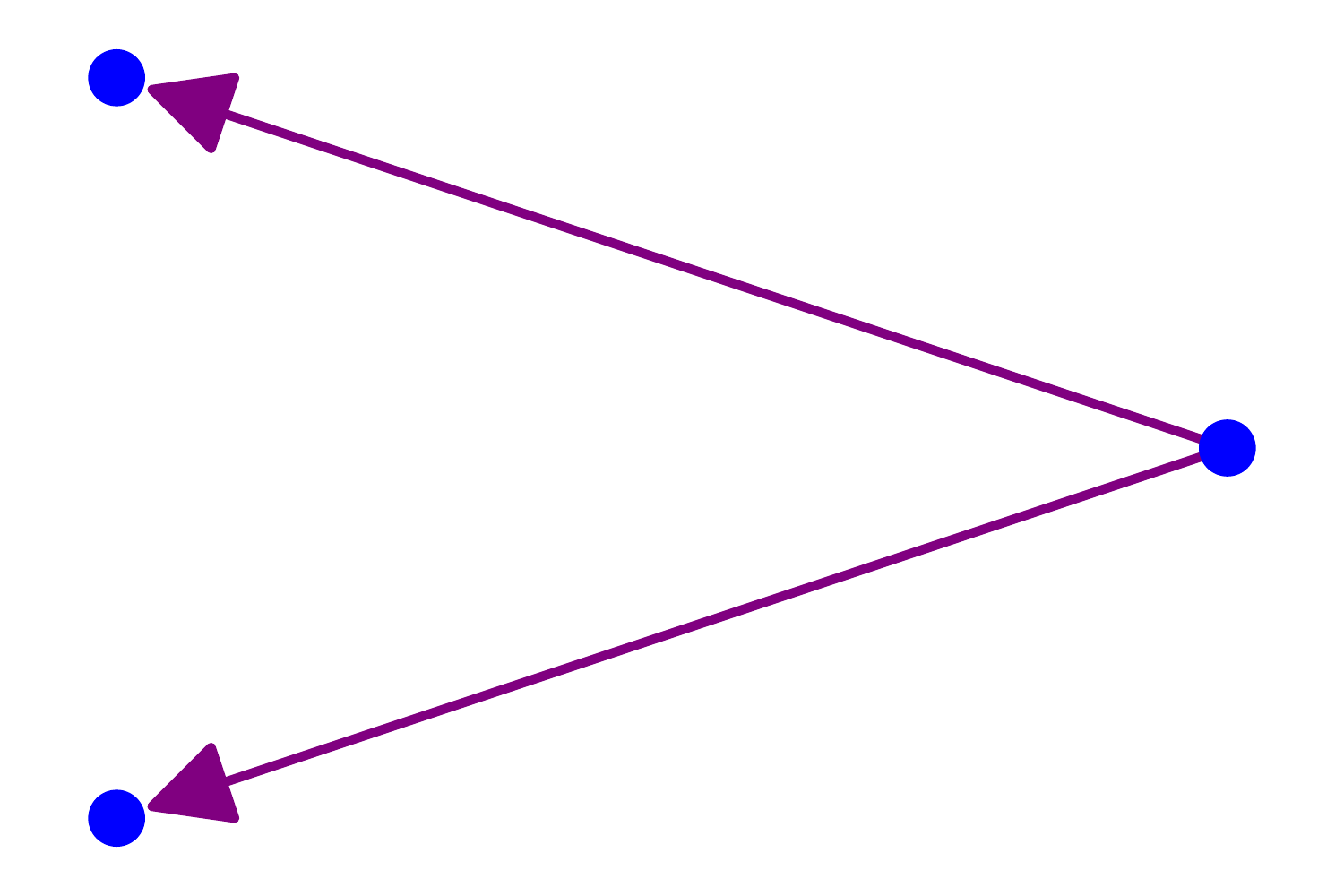}
    \caption{3-Fan out.}
    \label{fig:logistic2}
\end{subfigure}
\begin{subfigure}[b]{0.15\textwidth}
    \centering
    \includegraphics[width=\textwidth]{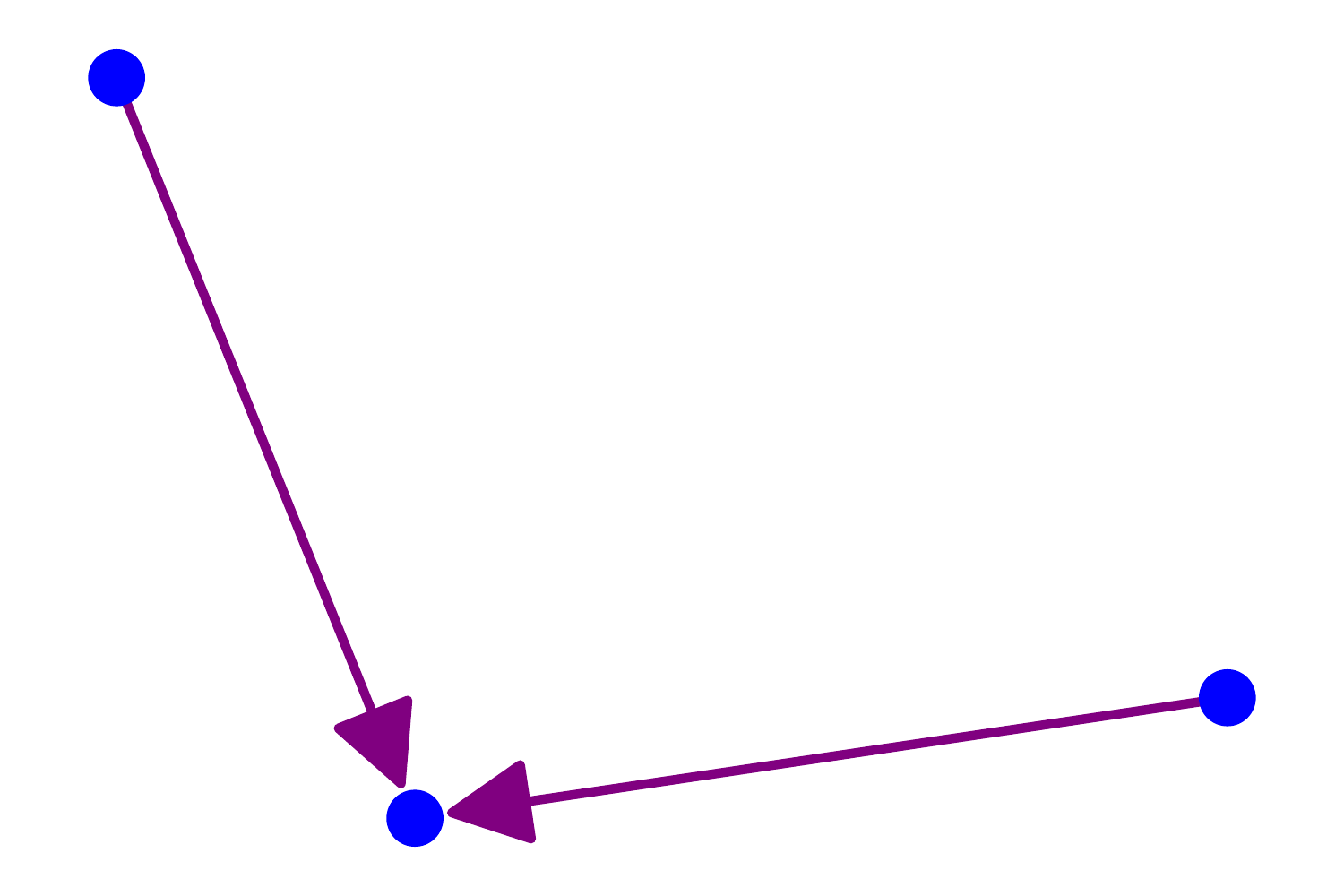}
    \caption{3-Fan in.}
    \label{fig:logistic3}
\end{subfigure}
\begin{subfigure}[b]{0.15\textwidth}
    \centering
    \includegraphics[width=\textwidth]{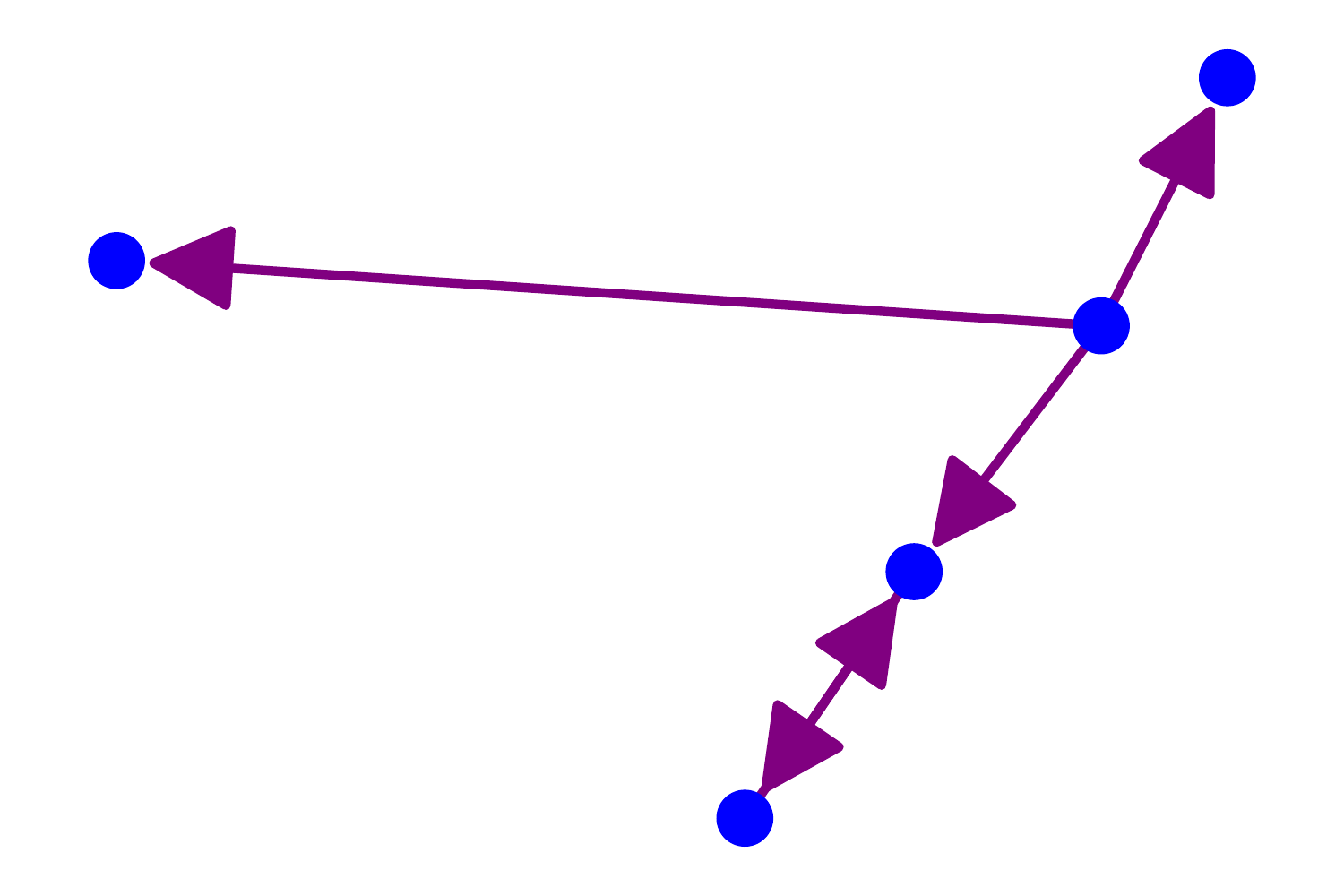}
    \caption{5-linear.}
    \label{fig:KGC_linear}
\end{subfigure}
\begin{subfigure}[b]{0.15\textwidth}
    \centering
    \includegraphics[width=\textwidth]{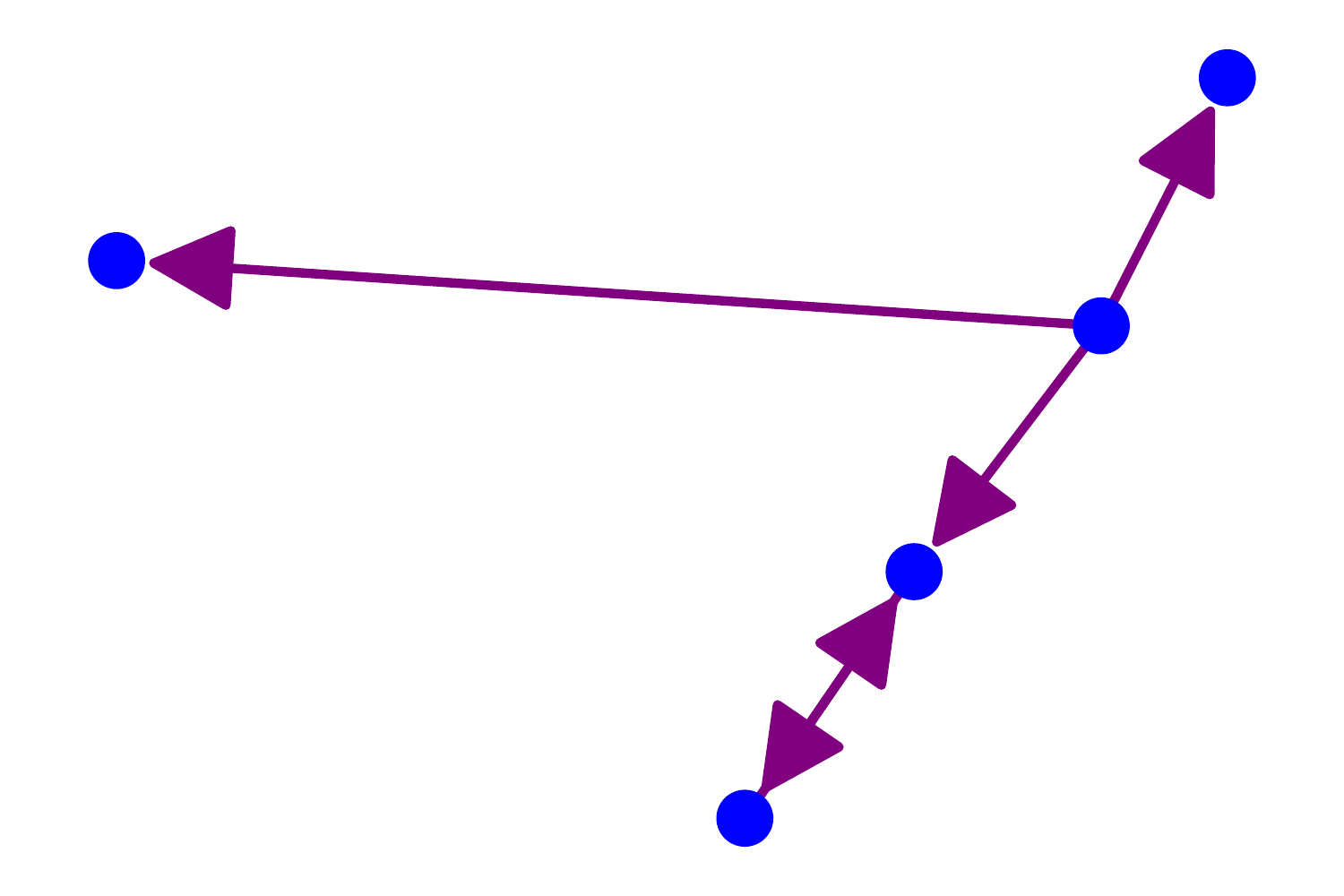}
    \caption{5-nonlinear.}
    \label{fig:KGC_nonlinear}
\end{subfigure}
        \caption{The topologies of datasets, which have a different number of nodes $\{2,3,5\}$, and different characteristics $\{\text{chaotic (a), linear (d), nonlinear (b, c, e)}\}$. Equations governing these networks are detailed in \cite{dsouza2020large}. }
        \label{fig:three graphs}
\end{figure}

\begin{figure*}[t]
    \centering
    \includegraphics[width=\linewidth]{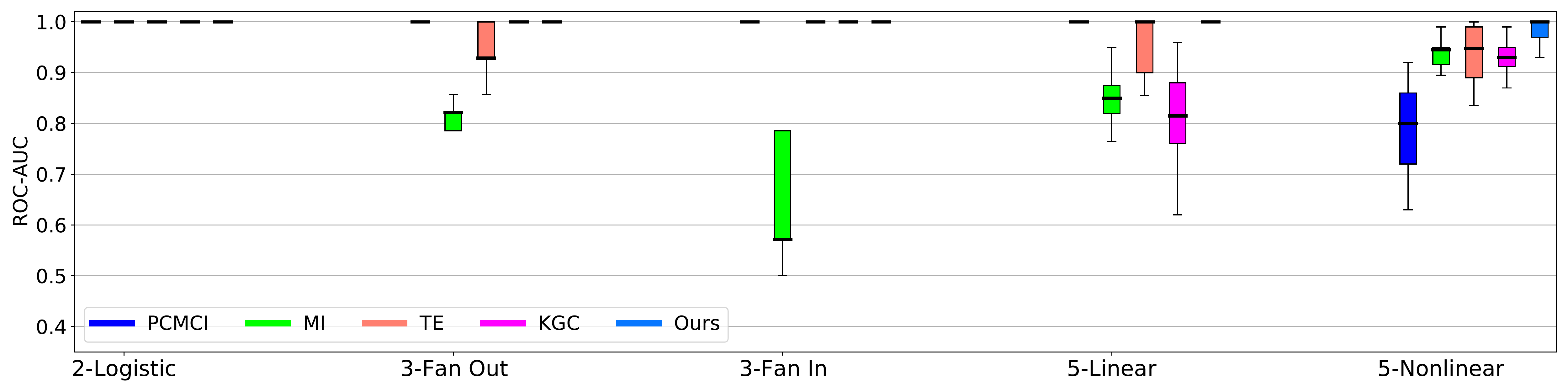}
     \caption{Five different datasets \textit{vs.} ROC-AUC of different methods are shown: a benchmark to compare PC-momentary conditional independence (PCMCI, 2019), mutual information (MI, 2018), transfer entropy (TE, 2018), kernel Granger causality (KGC, 2011) with our method. The number of time samples is T=500. Synthetic datasets with known ground truth and various topologies and dependencies (Fig. \ref{fig:three graphs}) each are replicated for 50 different random implementations. Our proposed method's performance (ROC-AUC) (light blue, first on the right side of each dataset) is significantly better than the competitor methods for all datasets. Boxplots represent [0.25,0.75] quartiles and median.}  
     \label{fig:7SYNTHETICS_AUC_boxplot}
     
\end{figure*}
\section{Numerical Tests}\label{sec:simuls}
Extensive simulations on a benchmark of the synthetic datasets with known ground-truth and a real-world dataset in a downstream task are presented in this section. 
\paragraph{Tests on synthetic datasets}
We used five various networks with various topologies, each with 50 random realizations, to test our method and some of the state-of-the-art methods, namely PCMCI (2019) \cite{runge2019detecting}, mutual information using Kraskov estimation (2018) \cite{wollstadt2018idtxl}, transfer entropy (TE) using Kraskov method (2018) \cite{kraskov2004estimating,wollstadt2018idtxl}, and kernel Granger causality 2011\cite{marinazzo2011nonlinear}. The topology of the datasets are shown in Fig. \ref{fig:three graphs}. We used one 2-nodes chaotic master-slave, two 3-nodes with fork and immorality topologies, and two 5-nodes datasets in the linear and nonlinear dependencies mode. Datasets are detailed in \cite{dsouza2020large}.

\begin{figure*}[t]
    \centering
    \includegraphics[width=0.95\linewidth]{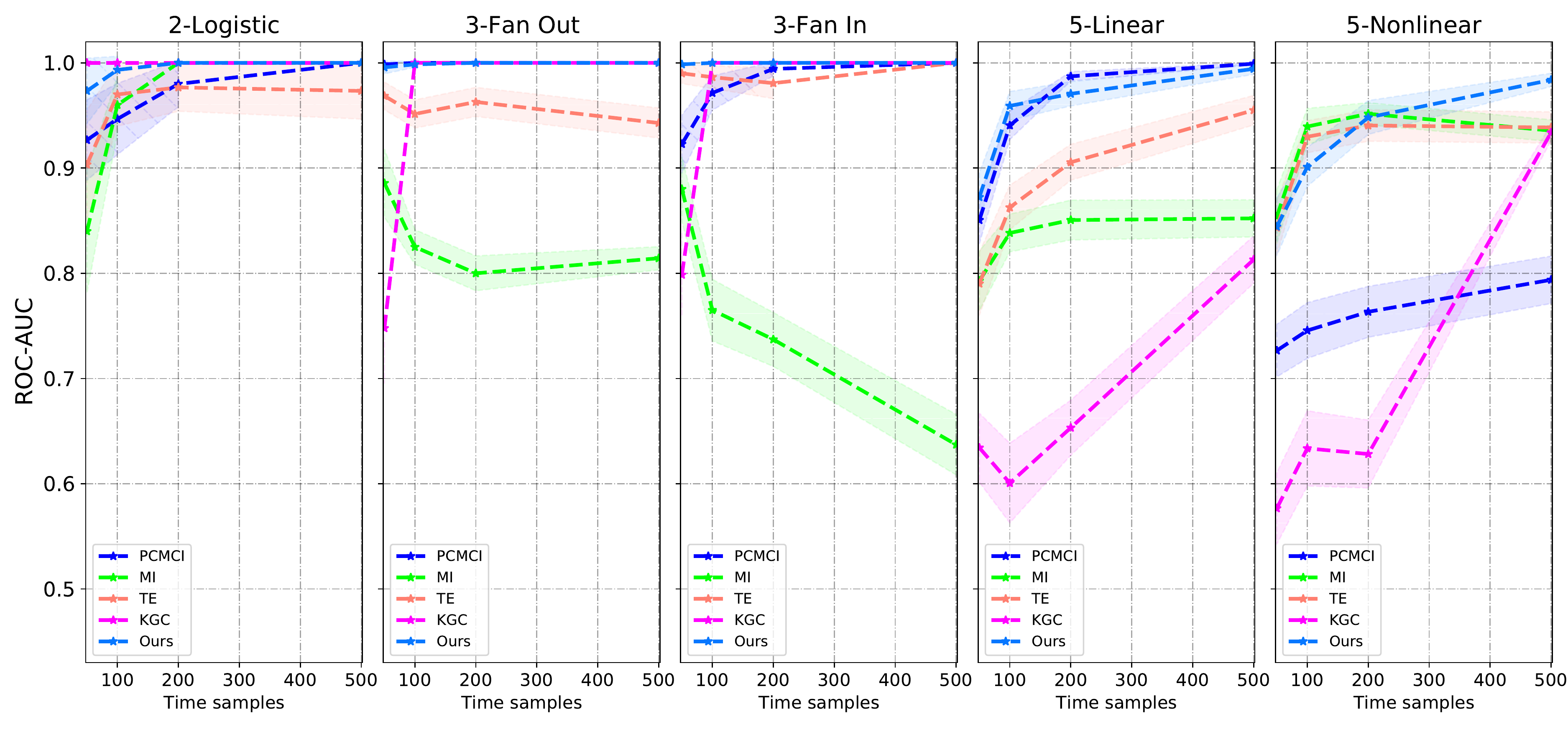}
     \caption{Simulations of various algorithms on different datasets are shown. Each column shows a different dataset (see titles). From left to right: 2-Logistic, 3-Fan out, 3-Fan in, 5-Linear, and 5-Nonlinear. Five different methods (PCMCI, MI, TE, KGC, and the proposed method) are tested for various time samples T= \{50, 100, 200, 500\} to compare the performance on the short time-series. Each dataset has 50 different replications. The confidence intervals of 95 percent are shown as shaded areas around the means. The time samples are increasing from left to right. Our proposed method (light blue, denoted by \textit{Ours}) is significantly more robust than the counterparts to the short time series and for all topologies.} 
     \label{fig:7SYNTHETICS_AUC_allTimes_boxplot}
     
\end{figure*}
We used datasets to address two challenges: 1) Fig. \ref{fig:7SYNTHETICS_AUC_boxplot} shows the case of which datasets have enough time samples for different topologies, and 2) Fig. \ref{fig:7SYNTHETICS_AUC_allTimes_boxplot} shows when the trend of different algorithms by reducing the number of time samples. ROC-AUC is used to compare the performance in the box plots. It is clearly seen in Fig. \ref{fig:7SYNTHETICS_AUC_boxplot} and \ref{fig:7SYNTHETICS_AUC_allTimes_boxplot} that our algorithm is more robust to the size of the network, nonlinearity in dependencies, and \textit{short time-series}. Note that PCMCI is also proposed for large-scale problems, but in our simulations, we show that our algorithm's robustness for short time-series is better than the PCMCI.

\begin{figure}
    \centering
    \includegraphics[width=.6\linewidth]{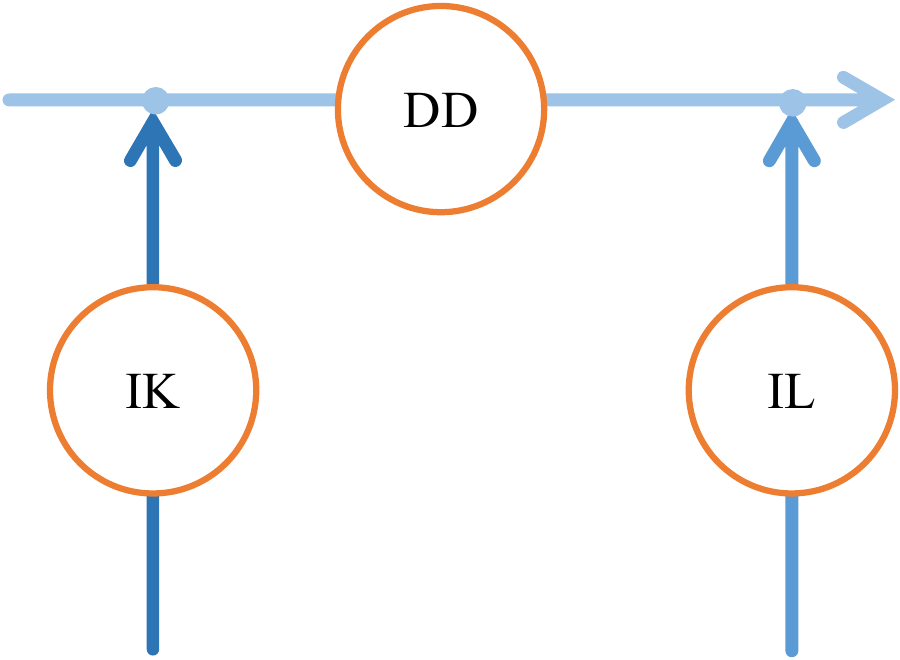}
    \caption{The average daily discharges of rivers in the upper Danube basin, the Iller at Kempten (IK), the Danube at Dillingen (DD), and the Isar at Lenggries (IL) are nonlinearly related. Only IK causes DD, which is correctly detected by the proposed method. }
    \label{fig:rivers_figure}
\end{figure}

\paragraph{Tests on real datasets} We examine our method on climatology data of average daily discharges of rivers in the upper Danube basin by three stations located on the Iller at Kempten (IK), the Danube at Dillingen (DD), and the Isar at Lenggries (IL). The data are available through Bavarian Environmental Agency at \textit{https://www.gkd.bayern.de}, and we use the measurements of three years (2017-2019). As shown in Fig. \ref{fig:rivers_figure}, there is a causal relationship IK $\xrightarrow{}$ DD since IK discharges into the DD upstream after a day, while there is no causal relationship between IL$\xrightarrow{}$IK and IL$\xrightarrow{}$DD (and \textit{vice versa}). Due to confounder variables such as rainfall or other weather conditions, the statistical analyses are vulnerable to detecting spurious connections.  Our method correctly detects a causal relationship between IK and DD while detecting no relationships between IL-IK and IL-DD, unconfounding spurious variables. 
In contrast, the Kraskov's TE \cite{wollstadt2018idtxl,schreiber2000measuring} wrongly detects the spurious connections as the DD$\xrightarrow{}$IK, DD$\xrightarrow{}$IL, IL$\xrightarrow{}$IK, and IL$\xrightarrow{}$DD altogether, for the best parameters. Our analysis of rivers' underlying dynamics conforms to  \cite{mhalla2020causal}, which shows our method's capability to distinguish between confounding and causal variables. 
\section{Conclusions} \label{sec:concs}
This paper puts forth a novel method for discovering nonlinear dependencies between time-series in multivariate environments. The method aims to solve curse-of-dimensionality while preserving nonlinear dependencies using combination of kernel principal component analysis, linear vector autoregressive model, and estimating pre-images to obtain topology of the graph in the input space. The method's advantage is its simplicity, while the results on the synthetic and real-world datasets show the merit of the method compared to state-of-the-art causality inference methods.


\bibliographystyle{IEEEtran}
\bibliography{refs}

\end{document}